\documentclass[twoside,leqno,twocolumn]{article}  
\usepackage{ltexpprt} 
\usepackage{times}
\usepackage{graphicx}
\usepackage{color}
\usepackage{multirow, multicol}
\usepackage{amsmath}
\usepackage{amssymb}
\usepackage{tikz}
\usepackage{longtable}
\usepackage{booktabs}
\usepackage{url}
\usepackage[sort,numbers]{natbib}

\newcommand{\nsv}{n_{\rm SV}}

\newcommand{\BLAS}{\texttt{BLAS}}
\newcommand{\ATLAS}{\texttt{ATLAS}}
\newcommand{\LOOPS}{\texttt{LOOPS}}
\newcommand{\expfun}{\sum_{i=1}^{n_{\rm SV}} \alpha_i y_i e^{-\gamma \| \mathbf{x}_i \|^2} e^{2\gamma \mathbf{x}_i^T\mathbf{z}}}

\newcommand{\affiliation}{KU Leuven, dept. of Electrical Engineering ESAT -- STADIUS \newline
KU Leuven -- iMinds dept. of Medical IT \newline
Kasteelpark Arenberg 10, box 2446; 3001 Leuven, Belgium}

\begin{document}

\title{\Large Fast Prediction with SVM Models Containing RBF Kernels\thanks{Research supported by: Research Council KUL:  ProMeta, GOA MaNet, KUL PFV/10/016 SymBioSys, START 1, OT 09/052 Biomarker, several PhD/postdoc $\&$ fellow grants; Flemish Government: IOF: IOF/HB/10/039 Logic Insulin, FWO: PhD/postdoc grants, projects: G.0871.12N research community MLDM; G.0733.09; G.0824.09; IWT: PhD Grants; TBM-IOTA3, TBM-Logic Insulin; FOD: Cancer plans; Hercules Stichting: Hercules III PacBio RS; EU-RTD: ERNSI; FP7-HEALTH CHeartED; COST: Action BM1104, Action BM1006: NGS Data analysis network; ERC AdG A-DATADRIVE-B.}}
\author{Marc Claesen\thanks{\affiliation} \\
\and Frank De Smet\thanks{KU Leuven, dept. of Public Health and Primary Care, Environment and Health; Kapucijnenvoer 35 blok d, box 7001; 3000 Leuven, Belgium} \\
\and Johan A.K. Suykens\footnotemark[2] \\
\and Bart De Moor\footnotemark[2]}
\date{}

\maketitle


\begin{abstract} \small\baselineskip=9pt We present an approximation scheme for support vector machine models that use an RBF kernel. A second-order Maclaurin series approximation is used for exponentials of inner products between support vectors and test instances. The approximation is applicable to all kernel methods featuring sums of kernel evaluations and makes no assumptions regarding data normalization. The prediction speed of approximated models no longer relates to the amount of support vectors but is quadratic in terms of the number of input dimensions. If the number of input dimensions is small compared to the amount of support vectors, the approximated model is significantly faster in prediction and has a smaller memory footprint. An optimized \verb-C++- implementation was made to assess the gain in prediction speed in a set of practical tests. We additionally provide a method to verify the approximation accuracy, prior to training models or during run-time, to ensure the loss in accuracy remains acceptable and within known bounds.
\end{abstract}


\section{Introduction}
Kernel methods form a popular class of machine learning techniques for various tasks. An important feature offered by kernel methods is the ability to model complex data through the use of the kernel trick \cite{scholkopf2002learning}. The kernel trick allows the use of linear algorithms to implicitly operate in a transformed feature space, resulting in an efficient method to construct models which are nonlinear in input space. In practice, despite the computationally attractive kernel trick, the prediction complexity of models using nonlinear kernels may prohibit their use in favor of faster, though less accurate, linear methods.

We present an approach to reduce the computational cost of evaluating predictive nonlinear models based on RBF kernels. This is valuable in situations where model evaluations must be performed in a limited time span. 
Several applications in the computer vision domain feature such requirements, including object detection \cite{cao:inria-00325810,10.1109/TPAMI.2012.62} and image denoising \cite{mika1999kernel,yang2004two}.

The widely used Radial Basis Function (RBF) kernel is known to perform well on a large variety of problems. It effectively maps the data onto an infinite-dimensional feature space. The RBF kernel function $\kappa(\cdot,\cdot)$ is defined as follows, with kernel parameter $\gamma$:
\begin{equation}
\kappa(\mathbf{x}_i,\mathbf{x}_j)=e^{-\gamma\|\mathbf{x}_i-\mathbf{x}_j\|^2}. \label{eq:rbf}
\end{equation}

Support vector machines (SVMs) are a prominent class of kernel methods for classification and regression problems \cite{burges1998tutorial}. The decision functions of SVMs take a similar form for various types of SVM models, including classifiers, regressors and least squares formulations \cite{Cortes:1995:SN:218919.218929,LSSVM}. For lexical convenience, we will use common SVM terminology in this text though the technique applies to all kernel methods.

The run-time complexity of kernel methods using an RBF kernel is $O(n_{SV}\times d)$ where $n_{SV}$ is the number of support vectors and $d$ is the input dimensionality. When run-time complexity is crucial and the number of support vectors is large, users are often forced to use linear methods which have $O(d)$ prediction complexity at the cost of reduced accuracy \cite{10.1109/TPAMI.2012.62}. We suggest a method which can significantly lower the run-time complexity of models with RBF kernels for many learning tasks.

In our approach, the decision function of SVM models that use an RBF kernel is approximated via the second-order Maclaurin series approximation of the exponential function. This approach was first proposed by Cao et al. \cite{cao:inria-00325810}. We extend their work by using fewer assumptions, providing a conservative bound on the approximation error for a given data set and reporting results of an extensive empirical analysis. Using this approximation, prediction speed can be increased significantly when the number of dimensions $d$ is low compared to the number of support vectors $n_{\rm SV}$ in a model. The proposed approximation is applicable to all models using an RBF kernel in popular SVM packages like LIBSVM \cite{CC01a}, SHOGUN \cite{SonRaeHenWidBehZieBonBinGehFra10} and LS-SVMlab \cite{lssvmlabguide}. 

We will derive the proposed approximation in the context of SVMs but its use easily extends to other kernel methods. Particularly, the approximation is applicable to all kernel methods that exploit the representer theorem \cite{scholkopf2001generalized}. This includes methods such as Gaussian processes \cite{rasmussen2006gaussian}, RBF networks \cite{poggio1990networks}, kernel clustering \cite{Filippone:2008:SKS:1284917.1285173}, kernel PCA \cite{scholkopf1998nonlinear, suykens2003support} and kernel discriminant analysis \cite{788121}.


\section{Related Work}
A large variety of methods exist to increase prediction speed. Three main classes of approaches can be identified: (i) pruning support vectors from models, (ii) approximating the feature space by a low-dimensional input space and (ii) approximating the decision function of a given model directly. Our proposed approach belongs to the latter class. 

\subsection{Reducing Model Size by Pruning Support Vectors}
Pruning support vectors linearly increases prediction speed because the run-time complexity of models with RBF kernels is proportional to the amount of support vectors. Pruning methods have been devised for SVM \cite{scholkopf1998fast,liang2010effective} and least squares SVM formulations \cite{suykens2002weighted,hoegaerts2004comparison}.

\subsection{Feature Space Approximations}
Rahimi and Recht proposed using standard linear methods after explicitly mapping the input data to a randomized low-dimensional feature space, which is designed such that the inner products therein approximate the inner products in feature space \cite{rahimi2007random}. This approach results in linear prediction complexity, as the resulting model is linear in the randomized input space. 
This is a general technique applicable to a large variety of kernel functions. For the RBF kernel, our specialized approach approximates each kernel evaluation to within $\epsilon = 0.03$ at complexity $O(d^2)$ when adhering to the proposed bounds. The complexity of random Fourier features is much higher than $O(d^2)$ for low-dimensional input spaces, where the RBF kernel is most useful \cite{rahimi2007random, cotter2011explicit}.

\subsection{Direct Decision Function Approximations}
Approaches that focus on approximating the decision function directly typically involve some form of approximation of the kernel function. Such approximations need not retain the structure and interpretation of the original model, provided that the decision function does not change significantly. Kernel approximations may leave out the interpretation of support vectors completely by reordering computations \cite{herbster2001learning}, or by aggregating support vectors into more efficient structures \cite{cao:inria-00325810}. Neural networks have also been used to approximate the SVM decision function directly \cite{Kang20144989}, in which case prediction speed depends on the chosen architecture.


A second-order approximation of the exponential function for RBF kernels was first introduced by Cao et al. \cite{cao:inria-00325810}. The basic concept of our paper resembles their work. In terms of training complexity, this approximation was analyzed in \cite{cotter2011explicit}. Here we focus exclusively on prediction speed. Cao et al. \cite{cao:inria-00325810} make two assumptions regarding normalization in deriving the approximations that may needlessly constrain their applicability. These assumptions are: 
\begin{enumerate}
\item Feature vectors are normalized to unit length, to simplify $\kappa$ to $\kappa(\mathbf{x}_i,\mathbf{x}_j)=e^{-2\gamma}e^{2\gamma\mathbf{x}_i^T\mathbf{x}_j}$.
\item Feature values must always be positive such that $0 \leq \mathbf{x}_i^T\mathbf{z} \leq 1$ holds.
\end{enumerate}
We will perform a more general derivation that requires none of these assumptions. Our derivation is agnostic to data normalization and we provide a conservative bound to assess the validity of the approximation during prediction (Eq.~\eqref{eq:bound}). Additionally, we derive the full approximation in matrix-form using the gradient and Hessian of the approximated part of the decision function. This allows the use of highly optimized linear algebra libraries in implementations of our work. Our benchmarks demonstrate that the use of such libraries yields a significant speed-up. Finally, we freely provide our implementation to facilitate comparison with competing approaches.


\section{Second-Order Maclaurin Approximation}

Predicting with SVMs involves computing a linear combination of inner products in feature space between the test instance $\mathbf{z} \in \mathbb{R}^{d}$ and all support vectors.  
In subsequent equations, $\mathbf{X} \in \mathbb{R}^{d \times \nsv}$ represents a matrix of $\nsv$ support vectors. 
We will denote the $i$-th support vector by $\mathbf{x}_i$ (the $i$-th column of $\mathbf{X}$). Via the representer theorem \cite{scholkopf2001generalized}, the decision values are a linear combination of kernel evaluations between the test instance and all support vectors:
\begin{equation}
f(\cdot)  : \mathbb{R}^d \rightarrow \mathbb{R} : f(\mathbf{z}) = \sum_{i=1}^{\nsv}\alpha_i y_i \kappa(\mathbf{x}_i,\mathbf{z}) \label{eq:generalcompact} + b,
\end{equation}
where $b$ is a bias term, $\alpha$ contains the support values, $\mathbf{y}$ contains the training labels and $\kappa(\cdot,\cdot)$ is the kernel function. 
Expanding the RBF kernel function \eqref{eq:rbf} in Eq.~\eqref{eq:generalcompact} yields:
\begin{align}
f(\mathbf{z}) &= \sum_{i=1}^{\nsv}\alpha_i y_i e^{-\gamma\|\mathbf{x}_i-\mathbf{z}\|^2} + b \nonumber \\
    &= \sum_{i=1}^{\nsv}\alpha_i y_i e^{-\gamma \|\mathbf{x}_i\|^2}e^{-\gamma \|\mathbf{z}\|^2} \underbrace{e^{2\gamma \mathbf{x}_i^T\mathbf{z}}} + b. \label{eq:general}
\end{align}

{\noindent}The exponentials of inner products between support vectors and the test instance -- underbraced in Equation~\eqref{eq:general} -- induce prediction complexity $O(\nsv\times d)$. Large models with many support vectors are slow in prediction, because each SV necessitates computing the exponential of an inner product in $d$ dimensions for every test instance $\mathbf{z}$. We use a second-order Maclaurin series approximation for these exponentials of inner products as described by \cite{cao:inria-00325810} (see the appendix for details on the Maclaurin series), which enables us to bypass the explicit computation of inner products.

{\noindent}The exponential per test instance $e^{-\gamma \|\mathbf{z}\|^2}$ can be computed exactly in $O(d)$. Before approximating the factors $e^{2\gamma \mathbf{x}_i^T\mathbf{z}}$, we reorder Equation~\eqref{eq:general} by moving the factor $e^{-\gamma \|\mathbf{z}\|^2}$ in front of the summation:
\begin{align}
f(\mathbf{z})&=e^{-\gamma \|\mathbf{z}\|^2} \big( \sum_{i=1}^{\nsv}\alpha_i y_i e^{-\gamma \|\mathbf{x}_i\|^2}e^{2\gamma \mathbf{x}_i^T\mathbf{z}} \big)+b, \nonumber \\
&=e^{-\gamma \|\mathbf{z}\|^2} g(\mathbf{z})+b, \label{eq:generalreordered}
\end{align}
with:
\begin{equation}
g(\cdot) : \mathbb{R}^d \rightarrow \mathbb{R} : g(\mathbf{z}) = \expfun. \label{eq:g}
\end{equation}

{\noindent}The exponentials of inner products can be replaced by the following approximation, based on the second-order Maclaurin series of the exponential function (see the appendix):
\begin{equation}
e^{2\gamma \mathbf{x}_i^T\mathbf{z}}\approx1+2\gamma \mathbf{x}_i^T\mathbf{z}+2\gamma^2(\mathbf{x}_i^T\mathbf{z})^2. \label{eq:approxexp}
\end{equation}

{\noindent}Approximating the exponentials $e^{2\gamma \mathbf{x}_i^T\mathbf{z}}$ in $g(\mathbf{z})$ \eqref{eq:g} via Equation~\eqref{eq:approxexp} yields:
\begin{align}
\hat{g}(\mathbf{z}) &= \sum_{i=1}^{\nsv}\alpha_i y_i e^{-\gamma \|\mathbf{x}_i\|^2}\big(1+2 \gamma \mathbf{x}_i^T\mathbf{z}+2 \gamma^2(\mathbf{x}_i^T\mathbf{z})^2\big) , \nonumber \\
&= c+\mathbf{v}^T\mathbf{z}+\mathbf{z}^T\mathbf{M}\mathbf{z}, \label{eq:ghat}
\end{align}
with:
\begin{align*}
c \in \mathbb{R} &= g(\mathbf{0}^{d}) = \sum_{i=1}^{\nsv}\alpha_i y_i e^{-\gamma \|\mathbf{x}_i\|^2}, \\
\mathbf{v} \in \mathbb{R}^{d} \rightarrow v_j &= \nabla g(\mathbf{z}) \\ 
&= 2 \gamma \sum_{i=1}^{\nsv}\alpha_i y_i e^{-\gamma \|\mathbf{x}_i\|^2} X_{j,i}, \\
\mathbf{v} &= \mathbf{X}\mathbf{w}, \\
\mathbf{M} \in \mathbb{R}^{d\times d} \rightarrow M_{j,k} &= \frac{\partial^2 g(\mathbf{z})}{\partial z_j \partial z_k} \\
&= 2 \gamma^2 \sum_{i=1}^{\nsv}\alpha_i y_i e^{-\gamma \|\mathbf{x}_i\|^2} X_{j,i} X_{k,i},\\
\mathbf{M} &= \mathbf{X}\mathbf{D}\mathbf{X}^T.
\end{align*}

{\noindent}The vector $\mathbf{v}$ and matrix $\mathbf{M}$ represent the gradient and Hessian of $g$, respectively. Here $\mathbf{w} \in \mathbb{R}^{\nsv}$ is a weighting vector: $w_i=2 \gamma \alpha_i y_i e^{-\gamma \|\mathbf{x}_i\|^2}$ and $\mathbf{D} \in \mathbb{R}^{\nsv\times \nsv}$ is a diagonal scaling matrix: $D_{i,i}=2 \gamma^2 \alpha_i y_i e^{-\gamma \|\mathbf{x}_i\|^2}$ and $D_{i,j}=0$ if $i\neq j$.  
Finally, the approximated decision function $\hat{f}(\mathbf{z})$ is obtained by using $\hat{g}(\mathbf{z})$ in Eq.~\eqref{eq:generalreordered}:
\begin{equation}
\hat{f}(\mathbf{z}) = e^{-\gamma \|\mathbf{z}\|^2} \hat{g}(\mathbf{z})+b = e^{-\gamma \|\mathbf{z}\|^2} \big( c+\mathbf{v}^T\mathbf{z}+\mathbf{z}^T\mathbf{M}\mathbf{z} \big) + b. \label{eq:approxcompact}
\end{equation}

{\noindent}The parameters $c$, $\mathbf{v}$, $\mathbf{M}$ and $b$ are independent of test points and need only be computed once. 
The complexity of a single prediction becomes $O(d^2)$ -- due to $\mathbf{z}^T\mathbf{Mz}$ -- instead of $O(\nsv\times d)$ for an exact RBF kernel.

The model size and prediction complexity of the proposed approximation is independent of the amount of support vectors in the exact model. This is especially interesting for least squares SVM formulations, which are generally not sparse in terms of support vectors \cite{LSSVM}. The RBF approximation loses its appeal when the number of input dimensions grows very large. For problems with high input dimensionality, the feature mapping induced by an RBF kernel often yields little improvement over using the linear kernel anyway \cite{hsu2003practical}.


\subsection{Approximation Accuracy}  \label{acc}
The relative error of the second-order Maclaurin series approximation of the exponential function is less than $3.05\%$ for exponents in the interval $[-0.5,0.5]$ (see Eq.~\eqref{eq:maclaurinbound} in Appendix~\ref{app:maclaurin}). Adhering to this interval guarantees that the relative error of any given term in the linear combination of $\hat{g}(\mathbf{z})$ is below $3.05\%$, compared to $g(\mathbf{z})$ (Eqs.~\eqref{eq:ghat} and \eqref{eq:g}, respectively). This translates into the following bound for our approximation:
\begin{equation}
|2\gamma \mathbf{x}_i^T \mathbf{z}| < \frac{1}{2}, \quad \forall i. \label{eq:bound1}
\end{equation}

{\noindent}The inner product can be avoided via the Cauchy-Schwarz inequality:
\begin{equation}
|\mathbf{x}_i^T\mathbf{z}| \leq \|\mathbf{x}_i\|\|\mathbf{z}\|, \quad \forall i. \label{eq:cauchy}
\end{equation}

{\noindent}Combining Eqs.~\eqref{eq:bound1} and \eqref{eq:cauchy} yields a way to assess the validity of the approximation in terms of the support vector $\mathbf{x}_M$ with maximal norm ($\forall i: \|\mathbf{x}_M \| \geq \| \mathbf{x}_i \|$):


\begin{equation}
 \|\mathbf{x}_M\|^2\|\mathbf{z}\|^2 < \frac{1}{16\gamma^2}. \label{eq:bound}
\end{equation}
Storing $\|\mathbf{x}_M\|^2$ in the approximated model enables checking adherence to the bound in Eq.~\eqref{eq:bound} during prediction, based on the squared norm of the test instance $\|\mathbf{z}\|^2$. Observe that this bound can be verified during prediction at no extra cost because $\|\mathbf{z}\|^2$ must be computed anyway (see Eq.~\eqref{eq:approxcompact}). 
Our tools can additionally report an upper bound for $\gamma$ for a given data set prior to training a model. In this case, the upper bound is obtained based on the maximum norm over all instances. The obtained upper bound for $\gamma$ may be slightly overconservative, because the data instance with maximum norm will not necessarily become a support vector.

\subsection{Relation to Degree-2 Polynomial Kernel}
The RBF approximation yields a quadratic form which can be related to a degree-2 polynomial kernel. We use the following general form for the degree-2 polynomial kernel:
\begin{equation}
\kappa(\mathbf{x}_i, \mathbf{x}_j) = \big(\gamma \mathbf{x}_i^T \mathbf{x}_j + \beta\big)^2.
\end{equation}
Note that $\gamma$ has a similar effect in the degree-2 polynomial kernel as in the RBF kernel (though not identical). To relate the second-order approximation of the RBF kernel with a degree-2 polynomial kernel we must expand the polynomial kernel in a similar fashion as in Equation~\eqref{eq:approxcompact}. Note that this expansion is exact for the polynomial kernel instead of an approximation as it is for the RBF kernel.

\newcounter{tempequationcounter}
\begin{figure*}[!t]
\normalsize
\setcounter{tempequationcounter}{\value{equation}}

\begin{align}
\textbf{approximated RBF} &\quad\longleftrightarrow\quad \textbf{exact degree-2 polynomial} \nonumber \\
{\color{blue}e^{-\gamma \|\mathbf{z}\|^2}} \big(c+\mathbf{w}^T\mathbf{X}\mathbf{z}+\mathbf{z}^T\mathbf{XDX}^T\mathbf{z}\big)+b 
    &\quad\longleftrightarrow\quad 
    c+\mathbf{w}^T\mathbf{X}\mathbf{z}+\mathbf{z}^T\mathbf{XDX}^T\mathbf{z} + b \label{eq:rbf-2d-decfun} \\
c = \sum_{i=1}^{\nsv}\alpha_i y_i {\color{blue}e^{-\gamma \|\mathbf{x}_i\|^2}} 
    &\quad\longleftrightarrow\quad
    c = \beta^2 \sum_{i=1}^{n_{SV}} \alpha_i y_i \label{eq:rbf-2d-c} \\
w_i = 2 \gamma \alpha_i y_i {\color{blue}e^{-\gamma \|\mathbf{x}_i\|^2}}
    &\quad\longleftrightarrow\quad
    w_i = 2 \beta \gamma \alpha_i y_i  \label{eq:rbf-2d-w} \\
D_{i,i} = {\color{blue}2} \gamma^2 \alpha_i y_i {\color{blue}e^{-\gamma \|\mathbf{x}_i\|^2}}
    &\quad\longleftrightarrow\quad 
    D_{i,i} = \gamma^2 \alpha_i y_i \label{eq:rbf-2d-d}
\end{align}
\setcounter{equation}{\value{tempequationcounter}}
\end{figure*}

Equations~\eqref{eq:rbf-2d-decfun} to \eqref{eq:rbf-2d-d} contrast an approximated RBF model with an exact model with degree-2 polynomial kernel. Fixing $\beta$ at $1$ facilitates the comparison which exposes two key differences between both models: (i) the nonlinearity $e^{-\gamma \|\mathbf{z}\|^2}$ in the approximated RBF model in Equation~\eqref{eq:rbf-2d-decfun} and (ii) a higher relative weight on second-order terms in the RBF approximation in Equation~\eqref{eq:rbf-2d-d}. The other exponential factors in terms of the support vectors in Eqs.~\eqref{eq:rbf-2d-c}-\eqref{eq:rbf-2d-d} act as scaling factors, which can be incorporated in the $\alpha$ values of the model with polynomial kernel to obtain an equivalent effect, e.g. $\alpha_i^{(2D)} = \alpha_i^{(RBF)} e^{-\gamma \|\mathbf{x}_i\|^2}$.

The extra scaling in Eq.~\eqref{eq:rbf-2d-decfun} adds flexibility to approximated RBF models compared to exact models with a polynomial kernel. The scaling causes the relative impact of the bias term $b$ in the model on the overall decision to vary per test instance $\mathbf{z}$. Adhering to the approximation bound defined in Equation~\eqref{eq:bound1} limits this scaling effect to the interval $(e^{-0.25}, 1]$, assuming $\|\mathbf{x}_M\| \geq \|\mathbf{z}\|,\ \forall \mathbf{z}$.


\subsection{Implementation} \label{implementation}
In order to benchmark the approximation against exact evaluations, we have made a \texttt{C++} implementation to approximate LIBSVM models and predict with the approximated model.\footnote{Our implementation is available at \url{https://github.com/claesenm/approxsvm}.} 
The implementation features a set of configurations to do the main computations. The configurations differ in the use of linear algebra libraries and vector instructions. Different configurations have consequences in two aspects: (i) approximating an SVM model and (ii) predicting with the approximated model.

\paragraph{Approximation Speed} The key determinant of approximation speed is matrix math. Approximation time is dominated by the computation of $\mathbf{M}=\mathbf{X}\mathbf{D}\mathbf{X}^T$, which involves large matrices if $d$ and $\nsv$ are large. The following implementations have been made:
\begin{enumerate}
\item \texttt{LOOPS}: uses simple loops to implement matrix math (default).
\item \texttt{BLAS}: uses the Basic Linear Algebra Subprograms (BLAS) for matrix math \cite{Blackford01anupdated}. The BLAS are usually available by default on modern Linux installations (in \texttt{libblas}). This default version is typically not heavily optimized.
\item \texttt{ATLAS}: uses the Automatically Tuned Linear Algebra Software (ATLAS) routines for matrix math \cite{Whaley00automatedempirical}. ATLAS provides highly optimized versions of the BLAS for the platform on which it is installed. The performance of ATLAS is comparable to vendor-specific linear algebra libraries such as Intel's Math Kernel Library \cite{eddelbuettel2010benchmarking}.
\end{enumerate}

\paragraph{Prediction Speed} The main factor in prediction speed for approximated models is evaluating $\mathbf{z}^T\mathbf{M}\mathbf{z}$ where $\mathbf{M}$ is a symmetric $d\times d$ matrix. This simple operation can exploit Single Instruction Multiple Data (SIMD) instruction sets if the platform supports them. The use of vector instructions can be enabled via compiler flags. We observed no significant gains in prediction speed when using the BLAS or ATLAS.


\section{Results and discussion} \label{classification}
To illustrate the speed and accuracy of the approximation, we used it for a set of classification problems. 
The exact models were always obtained using LIBSVM \cite{CC01a}. We investigated the amount of labels that differ between exact and approximated models as well as speed gains.

The accuracies are listed in Table~\ref{tab:accuracies}. We report the accuracy of the exact model and the percentage of labels which differ between the exact model and the approximation (note that not all differences are misclassifications). Table~\ref{tab:timings} reports the results of our speed measurements. Before discussing these results, we briefly summarize the data sets we used.

\subsection{Data Sets} 
To facilitate verification of our results, we used data sets that are freely available in LIBSVM format at the website of the LIBSVM authors.\footnote{Available at \url{http://www.csie.ntu.edu.tw/~cjlin/libsvmtools/datasets/}.} We used all the data sets as they are made available, without extra normalization or preprocessing. We used the following classification data sets:
\begin{itemize}
\item \texttt{a9a}: the Adult data set, predict who has a salary over $\$50.000$, based on various information \cite{Platt:1999:FTS:299094.299105}. This data set contains two classes, $d=123$ features (most are binary dummy variables) and $32,561$/$16,281$ training/testing instances.
\item \texttt{mnist}: handwritten digit recognition \cite{lecun1998gradient}. This data set contains 10 classes -- we classified class $1$ versus others, $d=780$ features and $60,000$/$10,000$ training/testing instances.
\item \texttt{ijcnn1}: used for the IJCNN 2001 neural network competition \cite{prokhorov2001ijcnn}. There are 2 classes, $d=22$ features and $49,990$/$91,701$ training/testing instances.
\item \texttt{sensit}: SensIT Vehicle (combined), vehicle classification \cite{duarte2004vehicle}. This data set contains 3 classes -- we classified class $3$ versus others, $d=100$ features and $78,823$/$19,705$ training/testing instances.
\item \texttt{epsilon}: used in the Pascal Large Scale Learning Challenge.\footnote{Available at \url{http://largescale.ml.tu-berlin.de/instructions/}.} This data set contains 2 classes, $d=2,000$ features and $400,000$/$100,000$ training/testing instances. To reduce training time, we switched the training and test set.
\end{itemize}

\subsection{Accuracy}
The accuracies we obtained in our benchmarks are listed in Table~\ref{tab:accuracies}. This table contains the key parameters per data set: number of dimensions $d$ and the maximum value that should be used for $\gamma$ in order to guarantee validity of the approximation ($\gamma_{MAX}$). Here $\gamma_{MAX}$ is obtained via Eq.~\eqref{eq:bound} after data normalization. The last column shows that only a very minor number of labels are predicted differently by the exact and approximated models.

Some of the listed results do not adhere to the bound, e.g. $\gamma > \gamma_{MAX}$. We used these parameters to illustrate that even though the accuracy of some terms in the linear combination may be inaccurate (e.g. relative error larger than $3\%$), the overall accuracy may still remain very good. In practice, we always recommend to adhere to the bound which guarantees high accuracy. Ignoring this bound is equivalent to abandoning all guarantees regarding approximation accuracy, because it is impossible to assess the approximation error which increases exponentially (shown in Figure~\ref{fig:maclaurinrelerr} in the appendix). 

When the bound was satisfied, the fraction of erroneous labels was consistently less than $1\%$ (\texttt{a9a}, \texttt{mnist} and \texttt{ijcnn1}). In the last experiment for \texttt{a9a} we used a value for $\gamma$ that is over five times larger than $\gamma_{MAX}$ and still get only $3.5\%$ of erroneous labels. These results demonstrate that the approximation is very acceptable in terms of accuracy.

The experiments on \texttt{sensit} and \texttt{epsilon} illustrate that a large number of dimensions $d$ safeguards the validity of the approximation to some extent, even when $\gamma$ becomes too large. The fraction $\gamma/\gamma_{MAX}$ is larger for \texttt{epsilon} than it is for \texttt{sensit} but due to the higher number of dimensions in \texttt{epsilon}, the fraction of erroneous labels remains lower ($0.53\%$ for \texttt{epsilon} versus $0.95\%$ for \texttt{sensit}). This occurs because the Cauchy-Schwarz inequality (Equation~\eqref{eq:cauchy}) is a worst-case upper bound for the inner product. When $d$ grows large, it is less likely for $|\mathbf{x}_i^T\mathbf{z}|$ to approach $\|\mathbf{x}_i\| \|\mathbf{z}\|$. In other words, the bound we use -- based on the Cauchy-Schwarz inequality -- is more conservative for larger input dimensionalities. 

\begin{table*}[t]
\centering
\begin{tabular}{lcccccccc}
\toprule 
data set & $d$ & $\gamma_{MAX}$ & $\gamma$ & $n_{\rm test}$ & $\nsv$ & $acc$ (\%) & $diff$ (\%) \\
\midrule 
adult (a9a) & $122$ & $0.018$ & $0.01$ & $16,281$ & $11,834$  & $84.8$ & $0.2$ \\
adult (a9a) & $122$ & $0.018$ & $0.02$ & $16,281$ & $11,674$ & $84.9$ & $1.3$ \\
adult (a9a) & $122$ & $0.018$ & $0.10$ & $16,281$ & $11,901$ & $85.0$ & $3.5$ \\
mnist & $780$ & $10^{-3}$ & $10^{-4}$ & $10,000$ & $2,174$ & $99.3$ & $0.08$ \\
ijcnn1 & $22$ & $0.064$ & $0.05$ & $91,701$ & $4,044$ & $97.7$ & $0.46$ \\
sensit & $100$ & $0.0025$ & $0.003$ & $19,705$ & $25,722$ & $86.5$ & $0.95$ \\
epsilon & $2000$ & $0.25$ & $0.35$ & $400,000$ & $36,988$ & $89.2$ & $0.53$ \\
\bottomrule
\end{tabular}
\caption{
Experiment summary: data set name, dimensionality, maximum value for $\gamma$, actual value for $\gamma$, number of testing points, number of SV in the model, accuracy of the (exact) model, number of differences between approximated and exact model.}
\label{tab:accuracies}
\end{table*}

\subsection{Speed Measurements}
Timings were performed on a desktop running Debian Wheezy. We used the default BLAS that are prebundled with Debian, which appear to be somewhat optimized, but not as much as ATLAS. We ran benchmarks on an Intel i5-3550K, which supports the Advanced Vector Extensions (AVX) instruction set for SIMD operations. 

Table~\ref{tab:timings} contains timing results of prediction speed between exact models and their approximations. The speed increase for the approximation is evident: ranging from $7$ to $137$ times when the time to approximate is disregarded, or $4.4$ to $69$ times when it is accounted for. We can see that the speed increase also holds for a large number of dimensions ($2000$ for the \texttt{epsilon} data set). The model for \texttt{mnist} contains few SVs compared to the number of dimensions, which causes a smaller speed increase in favor of the approximated model.

\begin{table*}[t]
\centering
\begin{tabular}{llcccccc}
\toprule
data set & approach & math & $t_{approx}$ (s) & SIMD & $t_{pred}$ (s) & ratio 1 & ratio 2 \\
\midrule
\texttt{a9a} & exact & $/$ & $/$ & $/$ &  $\mathbf{13.75\pm 0.060}$ & $1$ & $1$ \\
 & approx & \BLAS & $0.05\pm 0.002$ & $\times$ & $0.160\pm 0.002$ &  $86$ & $65$ \\
 &  & \LOOPS & $0.56\pm 0.021$ & $\checkmark$ & $0.146\pm 0.003$ & $94$ & $19$ \\
 & optimal & \BLAS & $0.05\pm 0.002$ & $\checkmark$ & $0.146\pm 0.003$ & $94$ & $70$ \\

\texttt{mnist} & exact & $/$ & $/$  & $/$ & $\mathbf{12.81\pm 0.016}$ & $1$ & $1$ \\
 & approx & \BLAS & $0.036\pm 0.001$ & $\times$ & $1.757\pm 0.008$ & $7.3$ & $7.1$ \\
 &  & \LOOPS & $1.480\pm 0.005$ & $\checkmark$ & $1.405\pm 0.006$ & $9.1$ & $4.4$ \\
 & optimal & \BLAS & $0.036\pm 0.001$ & $\checkmark$ & $1.405\pm 0.006$ & $9.1$ & $8.9$ \\

\texttt{ijcnn1} & exact & $/$ & $/$ & $/$ & $\mathbf{15.87\pm 0.012}$ & $1$ & $1$ \\
 & approx & \BLAS & $0.010\pm 0.000$ & $\times$ & $0.679\pm 0.012$ & $23$ & $23$ \\
 &  & \LOOPS & $0.010\pm 0.000$ & $\checkmark$ & $0.667\pm 0.016$ & $24$ & $23$ \\
 & optimal & \BLAS & $0.010\pm 0.000$ & $\checkmark$ & $0.667\pm 0.016$ & $24$ & $23$ \\

\texttt{sensit} & exact & $/$ & $/$ & $/$ & $\mathbf{79.62\pm 0.127}$ & $1$ & $1$ \\
 & approx & \BLAS & $0.670\pm 0.000$ & $\times$ & $0.590\pm 0.000$ & $134$ & $63$ \\
 &  & \LOOPS & $1.437\pm 0.036$ & $\checkmark$ & $0.581\pm 0.012$ & $137$ & $39$ \\
 & optimal & \ATLAS & $0.565\pm 0.005$ & $\checkmark$ & $0.581\pm 0.012$ & $137$ & $69$ \\

\texttt{epsilon} & exact & $/$ & $/$ & $/$ & $\mathbf{622.1\pm 0.165}$ & $1$ & $1$ \\
\texttt{*} & approx & \BLAS & $1.161\pm 0.003$ & $\times$ & $10.78\pm 0.110$ & $58$ & $52$ \\
 &  & \LOOPS & $43.98\pm 0.495$ & $\checkmark$ & $9.68\pm 0.03$ & $64$ & $12$ \\
 & optimal & \ATLAS & $0.442\pm 0.029$ & $\checkmark$ & $9.68\pm 0.03$ & $64$ & $61$ \\
\bottomrule
\end{tabular}
\caption{Comparison of prediction speed of an exact model vs. approximations. Approximations are classified based on use of math libraries and vector instructions. Times listed are prediction time and approximation time. Timings were performed in high-priority mode (using nice -3 in Linux) on an Intel i5-3550K at 3.30 GHz. ATLAS timings are not reported when its speed was comparable to BLAS. The last two columns contain the relative increase in prediction speed of the approximation compared to exact predictions, with and without accounting for the time needed to approximate the exact model.\newline \texttt{*}: time in minutes for the \texttt{epsilon} data set.} \label{tab:timings}
\end{table*}

In terms of approximation speed, the impact of specialized linear algebra libraries is apparant as shown in columns 3 and 4 of Table~\ref{tab:timings}. ATLAS consistently outperforms BLAS and both are orders of magnitude faster than the naive implementation, particularly when the matrix $\mathbf{X}$ gets large (over $100\times$ faster for \texttt{epsilon}, where $\mathbf{X}$ is $2.000\times 36.988$).

The impact of vector instructions is clear, with gains of up to $25\%$ in prediction speed when they are used (cfr. \texttt{mnist} results). Note that most of the time is spent on file IO for these benchmarks, which may give a pessimistic misinterpretation of the speed increase of vector instructions.

A competing method approximates the decision function using artificial neural networks (ANN) with a single hidden layer \cite{Kang20144989}. In this approach, prediction complexity is $O(n_{HN}\times d)$ where $n_{HN}$ is the number of hidden nodes in the network (typically $n_{HN} < n_{SV}$). \cite{Kang20144989} report prediction speedups of a factor $5$ to $28$ on models with few support vectors (which enables using few hidden nodes in the approximating ANN). The empirical speedup of using our quadratic approximation ranges from a factor $9$ to $137$ for models with many support vectors. When the number of support vectors grows, the decision boundary becomes more complex and will require more hidden units to be approximated effectively, which reduces the appeal of using ANNs. In contrast, the complexity of our approach is not influenced by the number of support vectors.



\section{Applications}
The most straightforward applications of the proposed approximation are those which require fast prediction. This includes many computer vision applications such as object detection, which require a large amount of predictions, potentially in real-time \cite{cao:inria-00325810,10.1109/TPAMI.2012.62}. 


Complementary to featuring faster prediction, the approximated kernel models are often smaller than exact models. The approximated models consist of three scalars ($b$, $c$ and $\gamma$), a dense vector ($\mathbf{v} \in \mathbb{R}^{d}$) and a dense, symmetric matrix ($\mathbf{M} \in \mathbb{R}^{d\times d}$). When the number of dimensions is small compared to the number of SVs, these approximated models are significantly smaller than their exact counterparts. We included Table~\ref{tab:modelsize} to illustrate this property: it shows the model sizes per classification data set. In our experiments the approximated models are smaller for all data sets except one. The biggest compression ratio we obtained was $290$ times (for the \texttt{sensit} data set). If we would approximate least squares SVM models, the compression ratios would be even larger due to the larger amount of SVs in least squares SVM models \cite{LSSVM}.

\begin{table}[!h]
\centering
\begin{tabular}{lccccc}
\toprule 
data set & $d$ & $\nsv$ & exact & approx & ratio \\
\midrule 
\texttt{a9a} & $122$ & $11,834$ & $830$ KB & $111$ KB & $7.5$ \\
\texttt{mnist} & $780$ & $2,174$ & $3.2$ MB & $3.7$ MB & $0.86$ \\
\texttt{ijcnn1} & $22$ & $4,044$ & $628$ KB & $4.2$ KB & $150$ \\
\texttt{sensit} & $100$ & $25,722$ & $32$ MB & $113$ KB & $290$ \\
\texttt{epsilon} & $2,000$ & $36,988$ & $1.1$ GB & $42$ MB & $27$ \\
\bottomrule
\end{tabular}
\caption{Comparison of model sizes (both types are stored in text format).}
\label{tab:modelsize}
\end{table}

Finally, a subtle side effect of our method is the fact that training data is obfuscated in approximated models. Data obfuscation is a technique used to hide sensitive data \cite{1366117}. Training data may be proprietary and/or contain sensitive information that cannot be exposed in contexts such as biomedical research \cite{murphy2002as}. In standard SVM models, the support vectors are exact instances of the training set. This renders SVM models unusable when data dissemination is an issue. The approximated models consist of complicated combinations of the support vectors (and typically $d \ll \nsv$), which makes it very challenging to reverse-engineer parts of the original data from the model. The approximation can be considered a surrogate one-way function of the support vectors \cite{DBLP:journals/corr/cs-CR-0012023}. As such, these approximations may allow SVMs to be used in situations where they are currently not considered \cite{1366117}.


\section*{Conclusion}

We have derived an approximation for SVM models with RBF kernels, based on the second-order Maclaurin series approximation of the exponential function. The applicability of the approximation is not limited to SVMs: it can be used in a wide variety of kernel methods. The proposed approximation has been shown to yield significant gains in prediction speed. 

Our benchmarks have shown that a minor loss in accuracy can result in very large gains in prediction speed. We have listed some example applications for such approximations. In addition to applications in which low run-time complexity is desirable, applications that require compact or data-hiding models benefit from our approach.

Our work generalizes the approximation proposed by \cite{cao:inria-00325810}. The derivation we performed made no implicit assumptions regarding data normalization. An easily verifiable bound was established which can be used to guarantee that the relative error of individual terms in the approximation remains low.

A competing method to approximate SVM models with an RBF kernel uses neural networks \cite{Kang20144989}. The advantages of our approach are (i) known bounds on the approximation error, (ii) faster to approximate an exact model (linear combination of SVs versus training a neural network) and (iii) faster in prediction when the number of dimensions is low. An advantage of the neural network approximation is that it can always be used, in contrast to our quadratic approximation whose validity depends on the data and choice of $\gamma$ as explained in Section~\ref{acc}.


\appendix
\section{Approximation of exponential function} \label{app:maclaurin}
The Maclaurin series for the exponential function and its second-order approximation are:
\begin{align}
e^x &= \sum_{k=0}^\infty \frac{1}{k!} x^k, \nonumber \\
 &\approx 1+x+\frac{1}{2}x^2. \label{eq:maclaurin}
\end{align}

\begin{figure}[!h]
  \centering
  \includegraphics[width=\linewidth]{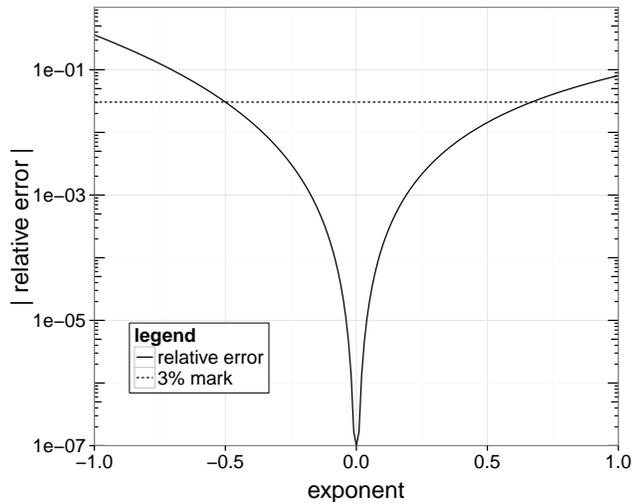}
  \caption{Absolute relative error of the second-order Maclaurin series approximation: $y(x) = \big|\big(e^x-(1+x+0.5x^2)\big)/e^x\big|$.} \label{fig:maclaurinrelerr}
\end{figure}

{\noindent}Figure~\ref{fig:maclaurinrelerr} illustrates the absolute relative error of the second-order Maclaurin series approximation. The relative error is smaller than $\pm 3\%$ when the absolute value of the exponent $x$ is small enough, e.g. $|x|<0.5$:
\begin{equation}
|x| < \frac{1}{2} \quad \Rightarrow \quad |\frac{e^x-1-x-0.5x^2}{e^x}| < 0.0305. \label{eq:maclaurinbound}
\end{equation}

{\noindent}This can be used to verify the validity of the approximation. 

\bibliographystyle{siam}
\bibliography{bibliography}

\end{document}